\documentclass[10pt,twocolumn,letterpaper]{article}
\usepackage{cvpr}
\usepackage{graphicx}
\usepackage{amsmath}
\usepackage{amssymb}
\usepackage{booktabs}
\usepackage{multirow}
\usepackage{calc}

\usepackage[pagebackref,colorlinks]{hyperref}
\usepackage[capitalize]{cleveref}
\usepackage[nolist]{acronym}
\usepackage[numbers]{natbib}
\usepackage{collect}
\usepackage{flushend}
\usepackage{adjustbox}
\usepackage{bm}
\usepackage{algorithm}
\usepackage{algpseudocode}
\usepackage{siunitx}

\DeclareMathOperator{\sgn}{sgn}

\def\cvprPaperID{5820}
\def\confName{CVPR}
\def\confYear{2022}

\usepackage{soul}
\usepackage{amsmath}
\usepackage{amssymb}
\usepackage{wrapfig}
\usepackage{booktabs}
\usepackage{multirow}
\usepackage{xspace}
\usepackage{xcolor}

\def\figurePath{images/}

\NewDocumentCommand{\rot}{O{45}
O{1em}m}{\makebox[#2][l]{\rotatebox{#1}{#3}}}%

\def\myfigure#1#2{%
    \begin{figure}[htb]%
    \centering\includegraphics*[width = \linewidth]{\figurePath#1}%
    \vspace{-.3cm}%
    \caption{#2}%
    \vspace{-.3cm}%
    \label{fig:#1}%
    \end{figure}%
}

\def\mycfigure#1#2{%
    \begin{figure*}[htb]%
    \centering\includegraphics*[width = \linewidth]{\figurePath#1}%
    \vspace{-.3cm}%
    \caption{#2}%
    \vspace{-.4cm}%
    \label{fig:#1}%
    \end{figure*}%
}

\newcommand{\argmin}[1]{\underset{#1}{\operatorname{arg\,min}\ }}

\newcommand{\refSec}[1]{Sec.~\ref{sec:#1}}
\newcommand{\refFig}[1]{Fig.~\ref{fig:#1}}
\newcommand{\refEq}[1]{Eq.~\ref{eq:#1}}
\newcommand{\refTab}[1]{Tab.~\ref{tab:#1}}
\newcommand{\refAlg}[1]{Alg.~\ref{alg:#1}}

\newcommand{\change}[1]{#1}

\soulregister\ref7
\soulregister\cite7
\soulregister\refFig7
\soulregister\refAlg7
\soulregister\cite7
\soulregister\ref7
\soulregister\pageref7
\soulregister\shortcite7
\soulregister\eg0
\soulregister\ie0
\soulregister\etal0

\DeclareGraphicsExtensions{.png,.jpg,.pdf,.ai,.psd}
\newcommand{\mysection}[2]{\vspace{-.1cm}\section{#1}\label{sec:#2}\vspace{-.1cm}}
\newcommand{\mysubsection}[2]{\subsection{#1}\label{sec:#2}\vspace{-.1cm}}

\definecollection{mymaths}
\newcommand{\mymath}[2]{
    \newcommand{#1}{\TextOrMath{$#2$\xspace}{#2}}
    \begin{collect}{mymaths}{}{}{}{}
    #1
    \end{collect}
}

\definecolor{colorA}{HTML}{4285f4}
\definecolor{colorB}{HTML}{ea4335}
\definecolor{colorC}{HTML}{fbbc04}
\definecolor{colorD}{HTML}{34a853}
\definecolor{colorE}{HTML}{ff6d01}
\definecolor{colorF}{HTML}{46bdc6}
\definecolor{colorG}{HTML}{000000}
\definecolor{colorH}{HTML}{777777}
\definecolor{colorI}{HTML}{bdd6ff}
\definecolor{colorJ}{HTML}{6a9e6f}

\usepackage{pifont}
\newcommand{\cmark}{\checkmark}%
\newcommand{\xmark}{\scalebox{0.85}{\ding{53}}}%

\newcommand{\expnum}[2]{$#1\!\!\times\!\!10^{#2}$}
\newcommand{\bexpnum}[2]{\boldsymbol{\expnum{#1}{#2}}}

\mymath{\parameters}{\theta}
\mymath{\parameterSpace}{\Theta}
\mymath{\scene}{f}
\mymath{\numberOfSamples}{N}
\mymath{\lightPath}{\mathbf{x}}
\mymath{\pixel}{P}
\mymath{\pathSpace}{\Omega}
\mymath{\diff}{\mathrm d}
\mymath{\noise}{Z}
\mymath{\objective}{\mathcal{L}}
\mymath{\kernel}{\kappa}
\mymath{\dimension}{\textrm{dim}}
\mymath{\pdf}{q}
\mymath{\offset}{\tau}
\newcommand{\estimate}[1]{\widehat{#1}}
\mymath{\cdf}{F}
\mymath{\bestparams}{\parameters^*}
\mymath{\refparams}{\parameters_{\textrm{ref}}}
\mymath{\initialparams}{\parameters_{\textrm{0}}}
\mymath{\currentparams}{\parameters_i}
\mymath{\perturbedMCsamples}{M}
\mymath{\perturbedImage}{Q}

\begin{document}

\begin{acronym}
\acro{AD}{Automatic Differentiation}
\acro{CDF}{Cumulative Distribution Function}
\acro{MC}{Monte Carlo}
\acro{GI}{Global Illumination}
\acro{PDF}{Probability Density Function}
\acro{NN}{Neural Network}
\acro{RE}{Rendering Equation}
\acro{MSE}{Mean Squared Error}
\acro{spp}{samples per pixel}
\acro{FD}{Finite Differences}
\acro{SPSA}{Simultaneous perturbation stochastic approximation}
\end{acronym}

\mymath{\density}{\sigma}

\newcommand{\task}[1]{\textsc{#1}}
\newcommand{\method}[1]{{\texttt{#1}}}
\newcommand{\mypara}[1]{\noindent\textbf{#1:}\quad}

\renewcommand{\eg}{\textit{e.g.}, }
\renewcommand{\ie}{\textit{i.e.}, }
\renewcommand{\wrt}{w.r.t.\ }

\newcommand{\supplTabParameters}{Tab. 1}
\newcommand{\supplFigOptim}{Fig. 1}

\title{Plateau-\change{reduced} Differentiable Path Tracing}

\author{Michael Fischer\\
University College London\\
{\tt\small m.fischer@cs.ucl.ac.uk}
\and
Tobias Ritschel\\
University College London\\
{\tt\small t.ritschel@ucl.ac.uk}
}
\maketitle

\begin{abstract}
Current differentiable renderers provide light transport gradients with respect to arbitrary scene parameters. 
However, the mere existence of these gradients does not guarantee useful update steps in an optimization. 
Instead, inverse rendering might not converge due to inherent plateaus, \ie regions of zero gradient, in the objective function.
We propose to alleviate this by convolving the high-dimensional rendering function, that maps scene parameters to images, with an additional kernel that blurs the parameter space.
We describe two Monte Carlo estimators to compute plateau-reduced gradients efficiently, \ie with low variance, and show that these translate into net-gains in optimization error and runtime performance. 
Our approach is a straightforward extension to both black-box and differentiable renderers and enables optimization of problems with intricate light transport, such as caustics or global illumination, that existing differentiable renderers do not converge on. Our code is at \url{github.com/mfischer-ucl/prdpt.}
\end{abstract}

\vspace{-0.4cm}
\mysection{Introduction}{Introduction}

Regressing scene parameters like object position, materials or lighting from 2D observations is a task of significant importance in graphics and vision, but also a hard, ill-posed problem.
When all rendering steps are differentiable, we can derive gradients of the final image \wrt the scene parameters. 
However, differentiating through the discontinuous rendering operator is not straightforward due to, \eg occlusion. 
The two main approaches to (differentiable) rendering are path tracing and rasterization.

Physically-based path-tracing solves the rendering equation by computing a \ac{MC} estimate for each pixel. 
Unfortunately, \ac{MC} is only compatible with modern \ac{AD} frameworks for the case of continuous integrands, \eg color, but not for spatial derivatives, \ie gradients \wrt an object's position.  
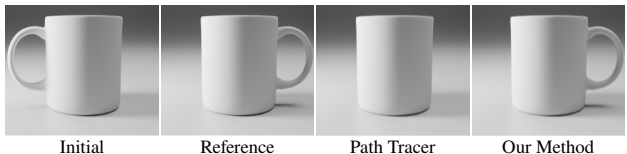
\myfigure{Teaser}{Optimization results with a differentiable path tracer (we use Mitsuba 3 \cite{nimier2019mitsuba}) and our proposed method. The task is to rotate the coffee cup around its $z$-axis, so that the handle moves to the right side. Due to a plateau in the objective function (when the handle is occluded by the cup), regular methods do not converge.}
To alleviate this, \citet{li2018differentiable} present re-sampling of silhouette edges and \citet{loubet2019reparameterizing} propose re-parametrizing the integrand, enabling the optimization of primitive- or light positions.
For rasterization, differentiability is achieved by replacing discontinuous edge- and $z$-tests with hand-crafted derivatives \cite{loper2014opendr,rhodin2015versatile,liu2019soft,le2021differentiable}.
The problem here is that rasterization, by design, does not capture complex light transport effects, \eg global illumination, scattering or caustics. 

Importantly, the mere existence of gradients is no guarantee that descending along them will make an optimization converge \cite{metz2021gradients}.
There are surprisingly many cases where they \textit{do not} lead to a successful optimization, due to a plateau in the objective function. 
An example is finding the orientation of the mug in \refFig{Teaser}:
As soon as the handle disappears behind the cup, no infinitesimally small rotation change will result in a reduced loss.
We have hence reached a plateau in the objective function, \ie a region of zero gradients.
We propose a method to remove these plateaus while still having complete, complex light transport.

We take inspiration from differentiable rasterization literature \cite{liu2019soft,rhodin2015versatile,petersen2022gendr,le2021differentiable}, where smoothing techniques are used to model the influence of faraway triangles to the pixel at hand.
For rasterization, this simple change has two effects: First, it makes the edge- and $z$-tests continuous and hence differentiable, and second, in passing (and, to our knowledge, much less studied), it also removes plateaus.
In this work, we hence aim to find a way to apply the same concept to complex light transport.
Therefore, instead of making the somewhat arbitrary choice of a fixed smoothing function for edge- and depth-tests in differentiable rasterizers, we path-trace an alternative, smooth version of the entire \ac{RE}, which we achieve by convolving the original \ac{RE} with a smoothing kernel. 
This leads to our proposed method, a lightweight extension to (differentiable) path tracers that extends the infinitely-dimensional path integral to the product space of paths and scene parameters.
The resulting double integral can still be \ac{MC}-solved efficiently, in particular with variance reduction techniques we derive (importance sampling and antithetic sampling). 

\mysection{Background}{RelatedWork}

\mysubsection{Rendering equation}{RenderingEquation}

According to the \ac{RE} \cite{kajiya1986rendering}, the pixel \pixel is defined as
\begin{equation}
    \label{eq:RE}
    \pixel(\parameters)
    = \int_{\pathSpace} \scene(\lightPath, \parameters)\diff\lightPath \,,
\end{equation}
an integral of the scene function $f(\lightPath, \parameters)$, that depends on scene parameters $\parameters\in\parameterSpace$, over all light paths $\lightPath\in\pathSpace$.
In \emph{inverse rendering}, we want to find the parameters \bestparams that best explain the pixels $\pixel_i$ in the reference image with

\begin{equation}
    \bestparams = 
    \argmin{\parameters} \, 
    \sum_i
    \objective \left( \pixel_i(\parameters) - \pixel_i(\refparams) \right)
    \,,
\end{equation}
where \objective is the objective function and $\pixel_i(\refparams)$ are the target pixels created by the (unknown) parameters \refparams.
Consider the example setting displayed in \refFig{sphere_examples}, where we are asked to optimize the 2D position of a circle so that its rendering matches the reference. 

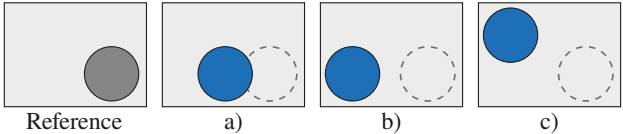
\myfigure{sphere_examples}{An example of a plateau in \objective: starting the optimization of the circle's position at a) will converge, whereas b) and c) will not. In a)-c), we show the reference dotted for convenience.}
Let \initialparams be the initial circle's position.
In this simple example, the optimization will converge if, and only if  the circle overlaps the reference, \ie setting a) in \refFig{sphere_examples}. 
The reason for this is that the gradient then is non-zero (a small change in \parameters is directly related to a change in \objective) and points towards the reference.
However, if there is no overlap between the initial circle and the reference, as in \refFig{sphere_examples} b), a gradient-based optimizer will not be able to recover the true position \refparams. 
This is due to the fact that there exists a \textit{plateau} in the objective function (for a rigorous definition, see \citet{jin2017escape}). 
To visualize this, consider a rendering where the circle is placed in the top left corner, as in \refFig{sphere_examples} c). 
The scalar produced by the objective function is identical for both b) and c), as \objective measures the distance in \textit{image space}. 
Therefore, the change in \objective is zero almost everywhere, leading to zero gradients and to the circle not moving towards the reference position.
As we will see in \refSec{Experiments}, this is surprisingly common in real applications. 

\newcommand{\methodsName}[1]{%
\multicolumn1c{%
\makebox[\widthof{Rasterize}][c]{#1}
}}
\begin{table}[htb]
    \setlength{\tabcolsep}{0.22cm}
    \centering
    \caption{Rendering taxonomoy. See \refSec{PathTracing} and \refSec{Rasterization}.}
    \label{tab:Methods}
    \begin{tabular}{lccc}
         &
         \methodsName{Rasterizer}&
         \methodsName{Path Tracer}&
         \methodsName{Ours}\\
         \toprule
         Differentiable & \cmark & \cmark & \cmark \\
         Light Transport & \xmark & \cmark & \cmark \\
         Plateau-reduced & \cmark & \xmark & \cmark \\
         \bottomrule
    \end{tabular}
\end{table}

\mysubsection{Path tracing}{PathTracing}
As there is no closed-form solution to \refEq{RE}, path tracing uses \ac{MC} to estimate the integral by sampling the integrand at random paths $\lightPath_i$:

\begin{equation}
    \estimate\pixel \approx \frac{1}{\numberOfSamples}\sum_i \scene(\lightPath_i, \parameters)
\end{equation}

\mypara{Gradients}
We are interested in the partial derivatives of \pixel with respect to the scene parameters \parameters, \ie

\begin{equation}
    \label{eq:derivativeintegral}
    \frac{\partial\pixel}{\partial\parameters} = 
    \frac{\partial}{\partial\parameters} \int_\pathSpace \scene(\lightPath, \theta)\diff\lightPath 
    = 
    \int_\pathSpace \frac{\partial}{\partial \parameters}\scene(\lightPath, \parameters)\diff\lightPath \,.
\end{equation}

In order to make \refEq{derivativeintegral} compatible with automatic differentiation, \citet{li2018differentiable} 
propose a re-sampling of silhouette edges and 
\citet{loubet2019reparameterizing} suggest a re-parametrization of the integrand. 
Both approaches allow to \ac{MC}-estimate the gradient as 

\begin{equation}
    \label{eq:MCgradient}
    \estimate{\frac{\partial\pixel}{\partial\parameters}}
    \approx
    \frac{1}{\numberOfSamples}
    \sum_i^\numberOfSamples
    \frac{\partial}{\partial \parameters}
    \scene(\lightPath_i, \parameters)
    \,.
\end{equation}

This is now standard practice in modern differentiable rendering packages \cite{nimier2019mitsuba, li2018differentiable, zeltner2021monte, zhang2021antithetic, zhang2020path}, none of which attempt to actively resolve plateaus.

\mysubsection{Rasterization}{Rasterization}
Rasterization solves a simplified version of the \ac{RE}, where for every pixel, the light path length is limited to one.
It is often used in practical applications due to its simplicity and efficiency, but lacks the ability to readily compute complex light transport phenomena.  
Instead, rasterization projects the primitives to screen space and then resolves occlusion.
\change{Both steps introduce jump discontinuities that, for differentiation, require special treatment.}

\mypara{Gradients}
In differentiable rasterization, both these operations \change{therefore} are replaced with smooth functions. 
\citet{loper2014opendr} approximate the spatial gradients during the backward pass by finite differences. 
Early, \citet{rhodin2015versatile}, often-used \citet{liu2019soft} and later \citet{petersen2019pix2vec} replace the discontinuous functions by soft approximations, \eg primitive edges are smoothened by the Sigmoid function. 
This results in a soft halfspace test that continuously changes \wrt the distance from the triangle edge and hence leads to a differentiable objective.
\citet{chen2019learning} and \citet{laine2020modular} propose more efficient versions of this, while
\change{\citet{xing2022differentiable} use an optimal-transport based loss function to resolve the problem.
However, most}
differentiable rasterizers make simplifying assumptions, \eg constant colors, the absence of shadows or reflections, and no illumination interaction between objects.
Our formulation does not make such assumptions. 

\mypara{Plateaus}
Choosing smoothing functions with infinite support (for instance, the Sigmoid), implicitly resolves the plateau problem as well.
Our method (\refSec{Method}) draws inspiration from this concept of ``differentiating through blurring''.

\mypara{Shortcomings} Consider again \refFig{sphere_examples} a), where the circle continuously influences the rendered image, resulting in a correct optimization outcome. 
For rasterizers, it is easy to construct a case where this does not hold, by imagining the circle to be the shadow of a sphere that is not seen in the image itself. 
The smoothed triangles then do not influence the rendering (most differentiable rasterizers do not even implement a shadow test \cite{li2018differentiable,petersen2022gendr}) and can therefore not be used for gradient computation.
Analogue examples can be constructed for other forms of multi-bounce light transport. 

\mysubsection{Other renderers}{Other}
Other ways to render that are neither path tracing or rasterization exist, such as differentiable volume rendering \cite{tulsiani2017multi,henzler2019escaping, mildenhall2021nerf} or fitting \acp{NN} to achieve a differentiable renderer \cite{nalbach2017deep,hermosilla2019deep,sanzenbacher2020learning, sitzmann2021light}.
Also very specific forms of light transport, such as shadows, were made differentiable \cite{lyu2021efficient}.
Finally, some work focuses on differentiable rendering of special primitives, such as points \cite{insafutdinov2018unsupervised,yifan2019differentiable}, spheres \cite{liu2020dist}, signed distance fields \cite{jiang2020sdfdiff, vicini2022differentiable, bangaru2022differentiable} or combinations \cite{cole2021differentiable, hu2022node}.
While some of these methods also blur the rendering equation and hence reduce plateaus, they remain limited to simple light transport.

\mysection{Plateau-reduced Gradients}{Method}

\mypara{Intuition}
As differentiable rasterization (cf. \refSec{Rasterization}) has established, the blurring of primitive edges is a viable means for differentiation.
But what if there is no ``primitive edge'' in the first place, as we deal with general light paths instead of simple triangles that are rasterized onto an image?
The edge of a shadow, for instance, is not optimizable itself, but the result of a  combination of light position, reflection, occlusion, etc.
Therefore, to achieve an effect similar to that of differentiable rasterizers, we would need a method that blurs the entire light path (instead of just primitive edges) over the parameter space \parameters.
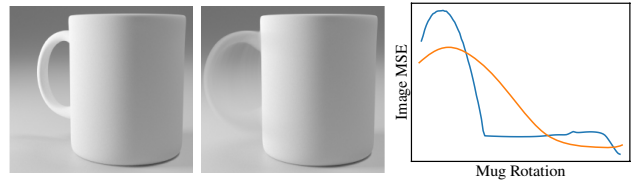
\myfigure{loss_cup_smoothed}{Optimizing the cup's rotation in the hard (left, blue) and smooth (right, orange) setting (note the blurred handle). The image-space loss landscape is displayed on the right: blurring resolves the plateau.}
If this method used a blur kernel with infinite support (\eg a Gaussian distribution), the plateau in the objective  would vanish, as a small parameter change in any direction would induce a change in the objective function.

\mypara{Example}
Let us consider \refFig{loss_cup_smoothed}, where we again want to optimize the cup's rotation around its $z$-axis to have the handle point to the right, a 1D problem. 
As we have seen previously, using an image-based objective function leads to a plateau when optimizing \objective in the ``hard'' setting, \ie without blur (the blue line in the plot).
Blurring the cup's rotation parameter, on the other hand, leads to \parameters continuously influencing the value of the objective and therefore resolves the plateau (orange line in the plot).
Naturally, it is easy to descend along the gradient of the orange curve, while the gradient is zero on the plateau of the blue curve.

\mysubsection{The Plateau-reduced Rendering Equation}{method_perturbedRE}

\mypara{Formulation}
We realize our blurring operation as a convolution of the rendering equation (\refEq{RE}) with a blur kernel \kernel over the parameter space \parameterSpace:
\begin{align}
    \label{eq:perturbedRE}
    \perturbedImage(\parameters) = 
    \kernel \star \pixel(\parameters)
    &=
    \int_{\parameterSpace\hphantom{\times \pathSpace}} \kernel(\offset)
    \int_\pathSpace \scene(\lightPath, \parameters - \offset) \, \diff\lightPath \diff\offset 
    \nonumber
    \\
    &=
    \int_{\parameterSpace \times \pathSpace}
    \kernel(\offset) 
    \scene(\lightPath, \parameters - \offset) \, 
    \diff\lightPath \diff\offset \,.
\end{align}
The kernel $\kernel(\offset)$ could be any symmetric monotonous decreasing function. 
For simplicity, we use a Gaussian here, but other kernels would be possible as well.
The kernel acts as a weighting function that weights the contribution of parameters \parameters that were offset by $\offset\in\parameterSpace$.
This means that, in addition to integrating all light paths \lightPath over \pathSpace, we now also integrate over all parameter offsets \offset in \parameterSpace.
We do not convolve across the path space \pathSpace but across the parameter space \parameters, \eg the cup's rotation in \refFig{loss_cup_smoothed}. 

\mypara{Estimation}
To estimate the (even higher-dimensional) integral in \refEq{perturbedRE}, we again make use of an \ac{MC} estimator

\begin{equation}
    \estimate\perturbedImage 
    \approx 
    \frac{1}{\numberOfSamples}
    \sum_i^\numberOfSamples
    \kernel(\offset) 
    \scene(\lightPath_i, \parameters-\offset_i) 
    \,,
\end{equation}

which is a practical approximation of \refEq{perturbedRE} that can be solved with standard path tracing, independent of the dimensionality of the light transport or the number of optimization dimensions. 

\mypara{Gradient }
Analogously to \refEq{MCgradient}, we can estimate the gradient of \perturbedImage through the gradient of its estimator
\begin{align}
    \label{eq:perturbedgradient}
    \estimate{
    \frac{\partial\perturbedImage}{\partial\parameters}
    }
    =
    \frac{\partial}{\partial\parameters}
    \frac{1}{\numberOfSamples}
    \sum_{i=1}^\numberOfSamples
    \kernel(\offset_i) 
    \scene(\lightPath_i, \parameters - \offset_i) 
    \,. 
\end{align}

Due to the linearity of differentiation and convolution, there are two ways of computing \refEq{perturbedgradient}: one for having a differentiable renderer, and one for a renderer that is not differentiable.
We discuss both options next.

\mypara{Plateau-reduced gradients if \pixel is differentiable}
With access to a differentiable renderer (\ie access to ${\partial\pixel} / {\partial \parameters}$), we can rewrite \refEq{perturbedgradient} as
\begin{align}
    \label{eq:PerturbedGradientDifferentiableRenderer}
    \estimate{
    \frac{\partial\perturbedImage}{\partial\parameters}
    }
    =
    \frac{1}{\numberOfSamples}
    \sum_{i=1}^\numberOfSamples
    \kernel(\offset_i)
    \underbrace{
    \frac{\partial\pixel}{\partial \parameters}
    (\parameters-\offset_i)
    }_\text{Diff. Renderer}
    \,.
\end{align}
Therefore, all that that needs to be done is to classically compute the gradients of a randomly perturbed rendering and weight them by the blur kernel.

\mypara{Plateau-reduced gradients if \pixel is not differentiable}
In several situations, we might not have access to a differentiable renderer, or a non-differentiable renderer might have advantages, such as computational efficiency, rendering features or compatibility with other infrastructure.
Our derivation also supports this case, as we can rewrite \refEq{perturbedgradient} as
\begin{align}  
    \label{eq:PerturbedGradientNonDifferentiableRenderer}
    \estimate{
    \frac{\partial\perturbedImage}{\partial\parameters}
    } 
    \approx    
    \frac{1}{\numberOfSamples}
    \sum_{i=1}^\numberOfSamples
    \underbrace{
    \frac{\partial\kernel}{\partial \parameters}
    (\offset_i)
    }_\text{Diff. Kernel}
    \underbrace{
    \vphantom{\frac{\partial}{\partial}}
    \pixel(\parameters-\offset_i)
    }_\text{Renderer}
    \,,
\end{align}
which equals sampling a non-differentiable renderer and weighting the result by the gradient of the blur kernel.
This is possible due to the additional convolution we introduce: prior work \cite{li2018differentiable, loubet2019reparameterizing} must take special care to compute derivatives (\refEq{MCgradient}), as in their case, optimizing \parameters might discontinuously change the pixel integral. 
We circumvent this problem through the convolution with \kernel, which ensures that, in expectation, \parameters \textit{continuously} influences the pixel integral. 

\mysubsection{Variance Reduction}{Variance reduction}
Drawing uniform samples from $\parameterSpace\times\pathSpace$ will result in a sample distribution that is not proportional to the integrand and hence lead to  high-variance gradient estimates and ultimately slow convergence for inverse rendering.
In our case, the integrand is the product of two functions (the kernel \kernel and the scene function \scene), which Veach  \cite{veach1998robust} showed how to optimally sample for.
As we generally consider the rendering operator a black box, we can only reduce variance by sampling according to the remaining function, the (differentiated) kernel \kernel (\refFig{ImportanceSampling}b).

\mymath{\bandwidth}{\sigma}
\mymath{\differentialKernel}{\nabla}
\mymath{\offsetComponent}{\offset_i}
\mymath{\bandwidthComponent}{\bandwidth_i}
\mymath{\uniformRandom}{\xi}
While importance-sampling for a Gaussian ($
\offset_i
\sim
\kernel
$, required to reduce variance of \refEq{PerturbedGradientDifferentiableRenderer}) is easily done, importance-sampling for the gradient of a Gaussian ($
\offset_i\sim
{\partial\kernel}/
{\partial \parameters}
$, to be applied to \refEq{PerturbedGradientNonDifferentiableRenderer}) is more involved.
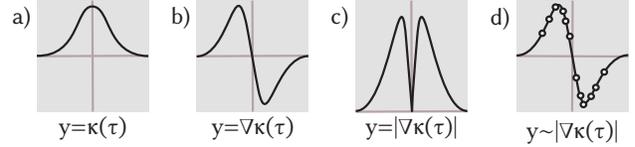
\myfigure{ImportanceSampling}{Our kernel \kernel (a), its gradient $\nabla\kernel$ (b), the positivized gradient (c) and samples drawn proportional thereto (d).}

The gradient of our kernel \kernel is 
\begin{equation}
\label{eq:gradientOfKernel}
\frac
{\partial\kernel}
{\partial \parameters}
(\offset)
=
\frac{-\offset}{\bandwidth^3\sqrt{2\pi}}
\exp
\left(
\frac
{-\offset^2}
{2\bandwidth^2}
\right)
\,,
\end{equation}
which is negative for $\offset>0$.
We enable sampling proportional to its \ac{PDF} by ``positivization'' \cite{owen2000safe}, and hence sample for $
|
\frac
{\partial\kernel}
{\partial \parameters}
(\offset)
|
$ 
instead (\refFig{ImportanceSampling}c).
We note that this function is separable at $\offset=0$ and thus treat each halfspace separately in all dimensions of \offset and \bandwidth. 
In order to sample we use the inverse \ac{CDF} method.
The \ac{CDF} of \refEq{gradientOfKernel} is
$$
\cdf(\offset) = 0.5 \sgn(\offset)
\exp\left(
-\frac{\offset^2}{2\bandwidth^2}\right) + C \,,
$$
where $C=1$ on the positive halfspace and 0 otherwise (this originates from the fact that the \ac{CDF} must be continuous, monotonically increasing and defined on $(0,1)$). 
Inverting the \ac{CDF} leads to
\begin{equation*}
    \cdf^{-1}(\uniformRandom)= \sqrt{-2\bandwidth^2\log(\uniformRandom)}\,,
\end{equation*}

into which we feed uniform random numbers $\uniformRandom\in(0,1)$ that generate samples proportional to $|\frac{\partial\kernel}{\partial \parameters}(\offset)|$ (\refFig{ImportanceSampling}d).

\change{Obtaining a zero-variance estimator for a positivized function requires sampling at at least two points: on the positivized and the regular part of the function \cite{owen2000safe}.}
We note that the function we sample for is point-symmetric around 0 in each dimension and hence use antithetic sampling \cite{hammersley1956general}, \ie for each sample $\offset$, we additionally generate its negated counterpart $-\offset$.
Doing so results in a zero-variance estimator, as we can perfectly sample for both parts of the function.
\change{In additional experiments, we found stratified sampling to be more brittle than antithetic sampling.}

Previous inverse rendering work applies antithetics to BSDF gradients in classic rendering
\cite{bangaru2020unbiased,zeltner2021monte} and to improve convergence on specular surfaces \cite{zhang2021antithetic}, while we use them as a means of reducing gradient variance for plateau-reduction, which is not present in such approaches.

\mymath{\bandwidthMin}{\bandwidth_m}
\mymath{\bandwidthInit}{\bandwidth_0}
\mysubsection{Adaptive bandwidth}{}
Adjusting \bandwidth gives us control over how far from the current parameter \parameters our samples will be spaced out. 
A high \bandwidth may be useful in the early stages of the optimization, when there still is a considerable difference between \parameters and \refparams, whereas we want a low \bandwidth towards the end of the optimization to zero-in on \refparams.
Throughout the optimization, we hence decay the initial \bandwidthInit according to a linear schedule, \ie $\bandwidth_{t+1} = \bandwidthInit - t (\bandwidthInit - \bandwidthMin)$, where \bandwidthMin is a fixed minimum value  we choose to avoid numerical instabilities that would otherwise arise from $\bandwidth \rightarrow 0$ in, \eg \refEq{gradientOfKernel}. \refFig{sphere_perturbed_progressive} shows the blur's progression throughout the optimization. 

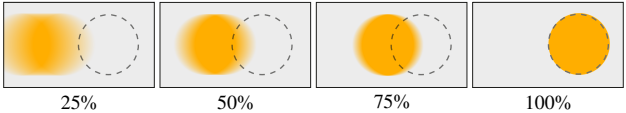
\myfigure{sphere_perturbed_progressive}{We visualize the adaptive spread of the smoothing at $n$\% of the optimization.
The reference position is shown dotted.}

\mysubsection{Implementation}{Implementation}
We outline our gradient estimator in pseudo-code in \refAlg{Method}.
We  importance-sample for our kernel with zero variance, use antithetic sampling and adapt the smoothing strength via \bandwidth.
As \refAlg{Method} shows, our method is simple to implement and can be incorporated into existing frameworks with only a few lines of code. 
We implement our method in PyTorch, with Mitsuba as rendering backbone, \change{and use Adam as our optimizer.}
For the remainder of this work, we use all components unless otherwise specified: importance sampling, adaptive smoothing and antithetic sampling. 
Moreover, we use \refEq{PerturbedGradientNonDifferentiableRenderer} for computational efficiency (cf. \refSec{Timing}) if not specified otherwise. 

\newcommand{\mycomment}[1]{\textcolor{gray}{\# #1}}
\begin{algorithm}[htb]
    \caption{Gradient estimation of the scene function \scene at parameters \parameters with perturbations \offset $\sim \mathcal{N}(0, \bandwidth)$ at \numberOfSamples samples.}
    \begin{algorithmic}[1]

    \State \mycomment{Equation 10}
    \Procedure{EstimateGradient}{%
        \pixel,
        \parameters,
        \bandwidth,
        \numberOfSamples
    }
        \State G := 0
        \For{$i\in(1, N/2)$}
        \State $\uniformRandom \leftarrow \Call{uniform}{0, 1}$
        \State $\offset \leftarrow \sqrt{-2\bandwidth^2 \log(\uniformRandom)}$
            
        \State $G \leftarrow 
            G + 
            \pixel(\parameters + \offset) \change{-}
            \pixel(\parameters - \offset)$
        \EndFor
        \State \Return G / \numberOfSamples
        \EndProcedure
    \end{algorithmic}
    \label{alg:Method}
\end{algorithm}

\mysection{Experiments}{Experiments}
We analyze our method and its variants in qualitative and quantitative comparisons against other methods and further compare their runtime performance. 
For the hyperparameters we use for our method and the competitors, please cf. the supplemental, \supplTabParameters.

\mysubsection{Methodology}{MethodsAndMetrics}
\newcommand{\methodGenDR}{\method{SoftRas}\xspace}
\newcommand{\methodMitsuba}{\method{Mitsuba}\xspace}
\newcommand{\methodOurGrad}{\method{Our$_{\kappa\partial\pixel}$}\xspace}
\newcommand{\methodOurNoGrad}{\method{Our$_{\partial\kappa\pixel}$}\xspace}

\mypara{Methods}
For our experiments, we compare four methods. 
The first is a differentiable rasterizer, SoftRas \cite{liu2019soft}.
Recall that soft rasterizers implicitly remove plateaus, which is why they are included here, despite their shortcomings for more complex forms of light transport.
For our method, we evaluate its two variants: The first uses a differentiable renderer and weights its gradients (\methodOurGrad, \refEq{PerturbedGradientDifferentiableRenderer}), while the second one performs differentiation through perturbation (\methodOurNoGrad, \refEq{PerturbedGradientNonDifferentiableRenderer}).
For both, we use Mitsuba 3 as our backbone, in the first variant using its differentiation capabilities to compute $\partial\pixel$, in the latter using it as a non-differentiable black-box to compute only \pixel. 
We run all methods for the same number of iterations and with the same rendering settings (\ac{spp}, resolution, path length, etc.).

\mypara{Metrics}
We measure the success of an optimization on two metrics, image- and parameter-space \ac{MSE}. 
As is common in inverse rendering, image-space \ac{MSE} is what the optimization will act on. 
Parameter-space \ac{MSE} is what we employ as a quality control metric during our evaluation.
This is necessary to interpret whether the optimization is working correctly once we have hit a plateau, as the image-space \ac{MSE} will not change there.
Note that we are not optimizing parameter-space \ac{MSE} and the optimization never has access to this metric.

\mypara{Tasks}
We evaluate our method and its competitors on six optimization tasks that feature advanced light transport, plateaus and ambiguities. We show a conceptual sketch of each task in \refFig{results_combined} and provide a textual explanation next. 

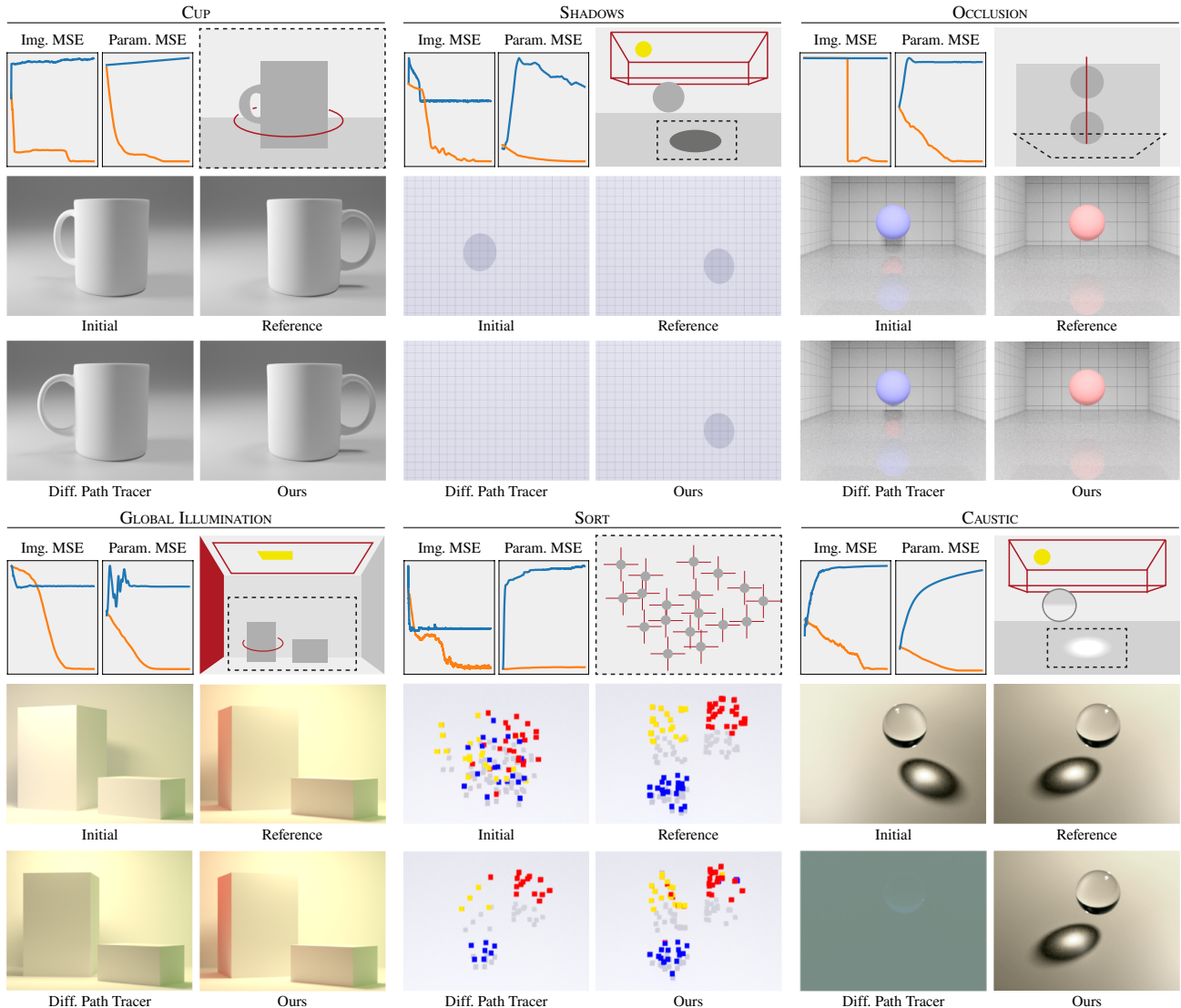
\mycfigure{results_combined}{We show the optimization tasks and results for \methodOurNoGrad (``Ours'', orange) and our baseline Mitsuba (``Diff. Path Tracer'', blue).}

\newcommand{\scinum}[2]{$#1$e{#2}}

\begin{table*}[t]
    \vspace{0.3cm}
    \centering
    \renewcommand{\tabcolsep}{0.15cm}    
    \caption{Image- and parameter-space \ac{MSE} of different methods (columns) on different tasks (rows).}
    \begin{tabular}{r rr rr rr rr}
    \toprule
        &
        \multicolumn2c{Rasterizer}&
        \multicolumn6c{Path Tracer}\\
        \cmidrule(lr){4-9}
        &
        \multicolumn2c{\methodGenDR}&
        \multicolumn2c{\methodMitsuba}&
        \multicolumn2c{\methodOurNoGrad}&
        \multicolumn2c{\methodOurGrad}
        \\
        \cmidrule(lr){2-3}
        \cmidrule(lr){4-5}
        \cmidrule(lr){6-7}
        \cmidrule(lr){8-9}
        &
        \multicolumn1c{\footnotesize{Img.}}&
        \multicolumn1c{\footnotesize{Para.}}&
        \multicolumn1c{\footnotesize{Img.}}&
        \multicolumn1c{\footnotesize{Para.}}&
        \multicolumn1c{\footnotesize{Img.}}&
        \multicolumn1c{\footnotesize{Para.}}&
        \multicolumn1c{\footnotesize{Img.}}&
        \multicolumn1c{\footnotesize{Para.}}
        \\
        \midrule
        \task{Cup}&
        \expnum{3.66}{-1} & 
        \expnum{2.72}{-2} & 
        \expnum{5.49}{-3} &
        \expnum{0.75}{-1} & 
        \bexpnum{4.92}{-6} &
        \bexpnum{4.18}{-7} & 
        \expnum{4.75}{-4} &
        \expnum{2.77}{-1}
        \\
        \task{Shadows}&
        \expnum{1.64}{-3} &
        \expnum{1.42}{-1} &
        \expnum{1.64}{-3} &
        \expnum{5.06}{-0} & 
        \bexpnum{1.74}{-5} & 
        \bexpnum{1.81}{-3} &
        \expnum{5.12}{-4} &
        \expnum{1.28}{-0}
        \\
        \task{Occl.}&
        \expnum{5.33}{-2} &
        \expnum{7.18}{-3} &
        \expnum{5.85}{-2} &
        \expnum{5.23}{+1} &
        \bexpnum{2.34}{-4} &
        \bexpnum{3.29}{-3} &
        \expnum{5.37}{-2} & 
        \expnum{1.87}{+1}
        \\
        \task{Global Ill.}&
        -- & 
        -- & 
        \expnum{3.78}{-2} &
        \expnum{3.87}{-1} &
        \bexpnum{5.07}{-5} &
        \bexpnum{8.71}{-4} &
        \expnum{5.88}{-2} &
        \expnum{2.55}{-1}
        \\
        \task{Sort}&
        \expnum{1.85}{-2} & 
        \expnum{1.57}{-0} & 
        \expnum{1.18}{-2} &
        \expnum{6.64}{-0} &
        \bexpnum{3.81}{-3} &
        \bexpnum{4.19}{-1} &
        \expnum{1.02}{-2} &
        \expnum{2.24}{-0}
        \\
        \task{Caustic}&
        -- &
        -- &
        \expnum{3.12}{-1} &
        \expnum{8.50}{-0} &
        \bexpnum{1.89}{-5} &
        \bexpnum{9.76}{-5} &
        \expnum{2.42}{-1} &
        \expnum{4.03}{-0}
        \\
    \bottomrule
    \end{tabular}
    \label{tab:Results}
   \vspace{-0.3cm}
\end{table*}

\mysubsection{Results}{results}
\mypara{Qualitative}
We now discuss our main result, \refFig{results_combined}.

\mypara{\task{Cup}} 
A mug is rotated around its vertical axis and as its handle gets occluded, the optimization has reached a plateau. 
Our method differentiates through the plateau. 
The differentiable path tracer gets stuck in the local minimum after slightly reducing the loss by turning the handle towards the left, due to the direction of the incoming light. 

\mypara{\task{Shadows}} An object outside of the view frustum is casting a shadow onto a plane. 
Our goal is to optimize the hidden object's position. 
Differentiable rasterizers can not solve this task, as they a) do not implement shadows, and b) cannot differentiate what they do not rasterize. 
The plateau in this task originates from the fact that the shadows do not overlap in the initial condition, which creates a situation akin to \refFig{sphere_examples} b). 
Again, our method matches the reference position very well.
Mitsuba first moves the sphere away from the plane (in negative $z$-direction), as this reduces the footprint of the sphere's shadow on the plane and thus leads to a lower error, and then finally moves the sphere out of the image, where a plateau is hit and the optimization can not recover.
The blue line in the image-space plot in \refFig{results_combined} illustrates this problem, as the parameter-error keeps changing very slightly, but the image-space error stays constant. 

\mypara{\task{Occlusion}}
Here, a sphere translates along the viewing axis to match the reference. 
The challenge is that the sphere initially is occluded by another sphere, \ie we are on a plateau as long as the occluder is closer to the camera than the sphere we are optimizing. 
The baseline path tracer initially pushes the red sphere towards the back of the box, as this a) reduces the error in the reflection on the bottom glass plane, and b) lets the shadow of the red sphere (visible underneath the blue sphere in the initial configuration) shrink, which again leads to a lower image-space error. 
Our method, in contrast, successfully differentiates through both the plateau (the red sphere has a negligible effect on the objective) and the discontinuity that arises when the red sphere first moves closer to the camera than the blue occluder. 

\mypara{\task{Global Illumination}} 
We here show that our method is compatible with the ambiguities encountered in advanced light transport scenarios. 
The goal of this optimization task is to simultaneously move the top-light to match the scene's illumination, change the left wall's color to create the color bleeding onto the box, and also to rotate the large box to an orientation where the wall's reflected light is actually visible. 
The optimization only sees an inset of the scene (as shown in \refFig{results_combined}; for a view of the full scene cf. the supplemental) and hence only ever sees the scattered light, but never the wall's color or light itself. 
The baseline cannot resolve the ambiguity between the box's rotation, the light position and the wall's color, as it is operating in a non-smoothed space. 
Our method finds the correct combination of rotation, light position and wall color.

\mypara{\task{Sort}} 
In this task, we need to sort a randomly perturbed assortment of 75 colored primitives into disjoint sets. 
We optimize the x- and y-coordinates of each cube, which leads to a 150-dimensional setting, with a plateau in each dimension, as most of the cubes are initially not overlapping their reference. 
Mitsuba cannot find the correct position of non-overlapping primitives and moves them around to minimize the image error, which is ultimately achieved by moving them outside of the view frustum. 
Our method, admittedly not perfect on this task, finds more correct positions, a result more similar to the reference. 

\mypara{\task{Caustic}} 
Lastly, the \task{Caustic} task features a light source outside the view frustum illuminating a glass sphere, which casts a caustic onto the ground. 
The goal is to optimize the light's position in order to match a reference caustic. 
As the sphere does not change its appearance with the light's movement, the optimization has to solely rely on the caustic's position to find the correct parameters. 
Similar to the \task{GI} task, this is not solvable for rasterizers. 
Our method  reaches the optimum position with high accuracy. 
For the baseline path tracer, we see a failure mode that is similar to the \task{Shadow} task. 
In this case, the image space error can be reduced by moving the light source far away, as the bulk of the error comes from the caustic not being cast onto the correct position. 
Moving the light source far away reduces this error, but also leads into a local minimum where there is no illumination at all, resulting in the gray image in \refFig{results_combined}. 

\mypara{Quantitative}
\refTab{Results} reports image- and parameter-space \ac{MSE} for all methods across all tasks. 
The quantitative results confirm what \refFig{results_combined} conveyed visually: regular gradient-based path tracers that operate on non-smooth loss landscapes fail catastrophically on our tasks. 
\methodGenDR manages to overcome some plateaus, but struggles with achieving low parameter error, as it blurs in image space but must compare to the non-blurred reference (as all methods), which leads to a notable difference between the final state and the reference parameters.   
To achieve comparable image-space errors, we render the parameters found by \methodGenDR with Mitsuba. 
Our method \methodOurNoGrad, in contrast, achieves errors of as low as $10^{-7}$, and consistently outperforms its competitors on all tasks by several orders of magnitude. 
Interestingly, \methodOurGrad (\ie using the gradients from the differentiable renderer) works notably worse than \methodOurNoGrad (but mostly still outperforms Mitsuba). 
We attribute this to the fact that we cannot importance-sample for the gradient here, as we do not know its \ac{PDF}. 
Instead, we can only draw samples proportional to the first term in the product, $\kernel(\offset)$, which places many samples where the kernel is high, \ie at the current parameter value. 
As we can see from the rigid optimization by Mitsuba, the gradient at the current parameter position is not very informative, so placing samples there is not very helpful.

\mysubsection{Timing}{Timing}

We now compare our approach's runtime and will see that, while \methodOurGrad needs more time to complete an optimization, \methodOurNoGrad on average is 8$\times$ faster than differentiable rendering with Mitsuba. 

Our method requires the additional step of (over-) sampling the parameter space in order to compute our smooth gradients. 
However, as shown in \refEq{PerturbedGradientNonDifferentiableRenderer}, our stochastic gradient estimation through the derivative-kernel (\methodOurNoGrad) allows us to skip the gradient computation step of the renderer.
While there exist techniques like the adjoint path formulation \cite{nimier2020radiative} and path replay backpropagation \cite{vicini2021path}, the gradient computation in inverse rendering still is computationally expensive and requires the creation of a gradient tape or compute graph.
Additionally, correct gradients \wrt visibility-induced discontinuities require a special integrator (re-parametrization or edge sampling). 

Our method \methodOurNoGrad, in contrast, does not need to compute $\partial\pixel/\partial \parameters$ and only requires a forward model. 
We can hence conveniently use the regular path integrator instead of its re-parametrized counterparts, and skip the gradient computation altogether. 
Moreover, our earlier efforts to develop an efficient importance-sampler will now pay off, as our method converges with as few as one extra sample only.
This results in notable speedups, and \methodOurNoGrad hence significantly outperforms other differentiable path tracers in wall-clock time at otherwise equal settings (cf. \refTab{timing}). 

\begin{table}[htb]
    \centering
    \renewcommand{\tabcolsep}{0.11cm}    
    \caption{Timing comparison for the three path tracing variants on all tasks. We report the average time for a single optimization iteration (with same hyperparameters) in seconds, so less is better.}
    \begin{tabular}{l r r r r r r}
    \toprule
        &
        \multicolumn1c{\task{Cup}}&
        \multicolumn1c{\task{Shad.}}&
        \multicolumn1c{\task{Occl.}}&
        \multicolumn1c{\task{GI}}&
        \multicolumn1c{\task{Sort}}&
        \multicolumn1c{\task{Caus.}}\\
        \midrule
        \methodMitsuba & 1.12\,s & 0.64\,s & 0.37\,s & 0.44\,s & 2.88\,s & 1.02\,s \\
        \methodOurNoGrad & \textbf{0.10\,s} & \textbf{0.04\,s} & \textbf{0.09\,s} & \textbf{0.15\,s} & \textbf{2.23\,s} & \textbf{0.08\,s} \\
        \methodOurGrad & 2.22\,s & 1.43\,s & 0.91\,s & 1.72\,s & 32.02\,s & 4.02\,s \\
    \bottomrule
    \end{tabular}
    \label{tab:timing}
\end{table}

\newcommand{\noIS}{\method{noIS}}
\newcommand{\noAP}{\method{noAP}}
\newcommand{\noAT}{\method{noAT}}

\mysubsection{Ablation}{ablation}
We now ablate our method to evaluate the effect its components have on the success of the optimization outcome. 
We will use \methodOurNoGrad from \refTab{Results} as the baseline and ablate the following components: importance sampling (\noIS), adaptive perturbations (\noAP) and antithetic sampling (\noAT). 
We hold all other parameters (\ac{spp}, resolution, etc.) fixed and run the same number of optimization iterations that was also used in \refTab{Results} and \refFig{results_combined}. 

We report the relative change between the ablations and our baseline in \refTab{Ablation}.
We report log-space values, as the results lie on very different scales.
From the averages in the last row, it becomes apparent that all components drastically contribute to the success of our method, while the most important part is the antithetic sampling. 
We emphasize that importance- and antithetic-sampling are \textit{variance reduction} techniques that do not bias the integration, \ie they do not change the integral's value in expectation. 
Therefore, our approach should converge to similar performance without these components, but it would take (much) longer, as the gradient estimates will exhibit more noise. 

\begin{table}[htb]
    \centering
    \renewcommand{\tabcolsep}{0.10cm}    
    \caption{Ablation of different components (columns) for different tasks (rows). We report the log-relative ratio \wrt \methodOurNoGrad, so higher values mean higher error.}
    \begin{tabular}{r rr rr rr}
    \toprule
        &
        \multicolumn2c{\noIS}&
        \multicolumn2c{\noAP}&
        \multicolumn2c{\noAT}
        \\
        \cmidrule(lr){2-3}
        \cmidrule(lr){4-5}
        \cmidrule(lr){6-7}
        &
        \multicolumn1c{\footnotesize{Img.}}&
        \multicolumn1c{\footnotesize{Para.}}&
        \multicolumn1c{\footnotesize{Img.}}&
        \multicolumn1c{\footnotesize{Para.}}&
        \multicolumn1c{\footnotesize{Img.}}&
        \multicolumn1c{\footnotesize{Para.}}
        \\
        \midrule
        \task{Cup}& 
        4.75$\times$&
        6.30$\times$&
        6.96$\times$&
        11.89$\times$&
        3.19$\times$&
        4.27$\times$
        \\
        \task{Shad.}& 
        5.27$\times$&
        5.98$\times$&
        3.07$\times$&
        1.27$\times$&
        5.30$\times$&
        6.03$\times$
        \\
        \task{Occl.}& 
        3.18$\times$&
        3.29$\times$&
        8.73$\times$&
        8.62$\times$&
        8.73$\times$&
        9.65$\times$
        \\
        \task{GI}& 
        10.38$\times$&
        10.84$\times$&
        6.40$\times$&
        9.15$\times$&
        5.62$\times$&
        12.06$\times$
        \\
        \task{Sort}& 
        1.48$\times$&
        0.70$\times$&
        1.14$\times$&
        2.04$\times$&
        1.59$\times$&
        1.09$\times$
        \\
        \task{Caus.}& 
        3.76$\times$&
        8.35$\times$&
        0.69$\times$&
        1.70$\times$&
        4.24$\times$&
        9.27$\times$
        \\
        \midrule
        Mean&
        4.81$\times$&
        5.91$\times$&
        4.50$\times$&
        5.78$\times$&
        4.78$\times$&
        7.06$\times$
        \\ 
    \bottomrule
    \end{tabular}
    \label{tab:Ablation}
\end{table}

\mysection{Discussion}{Discussion}

\paragraph{Related Approaches} Other methods proposed blurring by down-sampling in order to circumvent plateaus \cite{reddy2021im2vec,laine2020modular}. 
The quality upper bound for this is \methodGenDR, which we compare against in \refTab{Results}, as blurring by down-sampling does not account for occlusion, whereas \methodGenDR uses a smooth $z$-test. 
Another method that could be tempting to employ is \ac{FD}. 
Unfortunately, \ac{FD} does not scale to higher problem dimensions, as it requires $2n$ function evaluations on an $n$-dimensional problem 
\change{(on our \task{Sort} task, this would increase the per-iteration runtime by $\times 375$). 
A more economical variant is \ac{SPSA}, which perturbs all dimensions at once \cite{spall1992multivariate}. 
However, neither \ac{FD} nor \ac{SPSA} actively smoothes the loss landscape, as the gradient is always estimated from exactly two measurements, taken at fixed locations, often from a Bernoulli distribution.}
Our approach, in contrast, uses \numberOfSamples \emph{stochastic} samples, where \numberOfSamples is independent of the problem dimension.
In fact, we use $\numberOfSamples=2$ for most of our experiments (cf. Suppl. \supplTabParameters).
Our method's advantages thus are twofold: first, we do not require a fixed number or spacing of samples in the parameter space, but instead explore the space by stochastically sampling it. 
Second, our developed formalism allows to interpret this stochastic sampling as a means to compute a \ac{MC}-estimate of the gradient, and thus allows to simultaneously smooth the space and perform (smooth) differentiation. 

\change{Indeed, the formalism developed in \refSec{method_perturbedRE} can be interpreted as a form of variational optimization \cite{staines2012variational, staines2013optimization}, where one would descend along the (smooth) variational objective instead of the true underlying function.
As such, \refEq{PerturbedGradientNonDifferentiableRenderer} can be seen as an instance of a score-based gradient estimator \cite{sutton1999policy}, while \refEq{PerturbedGradientDifferentiableRenderer} can be interpreted as reparametrization gradient \cite{kingma2015variational, schulman2015gradient}. 
\citet{suh2022differentiable} provide intuition on each estimator's performance and align with our findings of the score-based estimator's superiority under a discontinuous objective.  
It is one of the contributions of this work to connect these variational approaches with inverse rendering.} 

\vspace{-0.15cm}
\paragraph{Limitations}
\change{As our method relies on Monte Carlo estimation, the variance increases favourably, but still increases with higher dimensions.
This can usually be mitigated by increasing the number of samples \numberOfSamples.
We show examples of a high-dimensional texture optimization in the supplemental.
Moreover, a good initial guess of \bandwidth is helpful for a successful optimization outcome (cf. Suppl. Sec. 2).
We recommend setting \bandwidth to roughly 50\,\% of the domain and fine-tune from there, if necessary.}

\mysection{Conclusion}{Conclusion}
We have proposed a method for inverse rendering that convolves the rendering equation with a smoothing kernel.
This has two important effects: it allows straight-forward differentiation and removes plateaus.
The idea combines strengths of differentiable rasterization and differentiable path tracing.
Extensions could include applying our proposed method to path tracing for volumes or Eikonal transport \cite{bemana2022eikonal,zhang2019differential} or other fields that suffer from noisy or non-smooth gradients, such as meta-learning for rendering \cite{fischer2022metappearance, liu2022learning}.
Our approach is simple to implement, efficient, has theoretical justification and optimizes tasks that existing differentiable renderers so far have diverged upon. 

\small{
\paragraph*{Acknowledgments}
This work was supported by Meta Reality Labs, Grant Nr. 5034015.
We thank Chen Liu, Valentin Deschaintre and
the anonymous reviewers for their insightful feedback.
}

\setlength{\bibsep}{0.0pt}
{\small
\bibliographystyle{plainnat}
\bibliography{main}
}

\end{document}


\begin{acronym}
\acro{AD}{Automatic Differentiation}
\acro{CDF}{Cumulative Distribution Function}
\acro{MC}{Monte Carlo}
\acro{GI}{Global Illumination}
\acro{PDF}{Probability Density Function}
\acro{NN}{Neural Network}
\acro{RE}{Rendering Equation}
\acro{MSE}{Mean Squared Error}
\acro{spp}{samples per pixel}
\acro{FD}{Finite Differences}
\acro{ICDF}{Inverse \ac{CDF}}
\acro{FOV}{Field of View}
\end{acronym}

\mymath{\parameters}{\theta}
\mymath{\parameterSpace}{\Theta}
\mymath{\scene}{f}
\mymath{\numberOfSamples}{N}
\mymath{\lightPath}{\mathbf{x}}
\mymath{\pixel}{P}
\mymath{\pathSpace}{\Omega}
\mymath{\diff}{\mathrm d}
\mymath{\noise}{Z}
\mymath{\objective}{\mathcal{L}}
\mymath{\kernel}{\kappa}
\mymath{\dimension}{\textrm{dim}}
\mymath{\pdf}{q}
\mymath{\offset}{\tau}
\newcommand{\estimate}[1]{\widehat{#1}}
\mymath{\cdf}{F}

\mymath{\bestparams}{\parameters^*}
\mymath{\refparams}{\parameters_{\textrm{ref}}}
\mymath{\initialparams}{\parameters_{\textrm{0}}}
\mymath{\currentparams}{\parameters_i}

\mymath{\perturbedMCsamples}{M}
\mymath{\perturbedImage}{Q}

\mymath{\bandwidth}{\sigma}
\mymath{\differentialKernel}{\nabla}
\mymath{\offsetComponent}{\offset_i}
\mymath{\bandwidthComponent}{\bandwidth_i}
\mymath{\uniformRandom}{\xi}
\mymath{\bandwidthMin}{\bandwidth_m}
\mymath{\bandwidthInit}{\bandwidth_0}
\mymath{\warmstartIterations}{w}
\mymath{\iterations}{k}
\mymath{\spp}{\textrm{spp}}

\newcommand{\task}[1]{\textsc{#1}}
\newcommand{\method}[1]{{\texttt{#1}}}
\newcommand{\mypara}[1]{\noindent\textbf{#1:}\quad}

\renewcommand{\eg}{\textit{e.g.}, }
\renewcommand{\ie}{\textit{i.e.}, }
\renewcommand{\wrt}{\textit{w.r.t.} }

\newcommand{\supplTabParameters}{Tab. 1}
\newcommand{\supplFigOptim}{Fig. 1}

\newcommand{\mainTabResults}{Tab. 2\xspace}
\newcommand{\mainTabTiming}{Tab. 3\xspace}
\newcommand{\mainPseudocode}{Alg. 1\xspace}
\newcommand{\mainGradientEquation}{Eq. 11\xspace}

\newcommand{\methodGenDR}{\method{SoftRas}\xspace}
\newcommand{\methodMitsuba}{\method{Mitsuba}\xspace}
\newcommand{\methodOurGrad}{\method{Our$_{\kappa\partial\pixel}$}\xspace}
\newcommand{\methodOurNoGrad}{\method{Our$_{\partial\kappa\pixel}$}\xspace}

\newcommand{\methodRedner}{\method{Redner}\xspace}

\title{Plateau-reduced Differentiable Path Tracing: Supplemental}

\author{Michael Fischer\\
University College London\\
{\tt\small m.fischer@cs.ucl.ac.uk}
\and
Tobias Ritschel\\
University College London\\
{\tt\small t.ritschel@ucl.ac.uk}
}
\maketitle

This supplemental contains the hyperparameters we used for our experiments (\refSec{hyperparameters}), including an additional analysis of our two main parameters \numberOfSamples and \bandwidth (\refSec{analysis}), experiments on compatibility with plateau-free problems and other renderers (\refSec{compatibility}) and the derivations of the equations presented in the main text (\refSec{derivations}). 

\mysection{Hyperparameters}{hyperparameters}

\mypara{Hyperparameters} \refTab{hyperparams} shows all the hyperparameters we use for our main experiments for all tasks. The first two columns are hyperparameters of our approach:
\numberOfSamples is the number of samples we use during an optimization iteration (for an analysis, cf. \refFig{n_vs_error}), and \bandwidthInit is the kernel spread with which we start the optimization (for an analysis, cf. \refFig{sigma_vs_error}).
\Ac{spp} is the rendering setting we use for rendering with Mitsuba, which we generally didn't tune and hence don't regard as a hyperparameter of our method, but set such that the noise is less than the signal we want to optimize. 
We use the same \ac{spp} across all path-tracing methods.
LR is the optimizer's learning rate (we use Adam with default parameters) and the last column shows the number of optimization iterations we run. 
We warm-start our \bandwidth annealing schedule after approx. 50\% of the optimization and use $\bandwidthMin=0.01$ for all experiments as the lowest value we decrease \bandwidth to during the annealing schedule, in order to avoid numerical instabilities. 

\begin{table}[htb]
    \centering
    \caption{Experiment parameters (columns) for all tasks (rows).}
    \renewcommand{\tabcolsep}{0.35cm}    
    \begin{tabular}{l rrrrr}
    \toprule
        &
        \multicolumn1c{\numberOfSamples}&
        \multicolumn1c{\bandwidthInit}&
        \multicolumn1c{\spp}&
        \multicolumn1c{LR}& 
        \multicolumn1c{Iter.} 
        \\
        \midrule
        \task{Cup} & 2 & 0.250 & 16 & 0.01 &  400 \\
        \task{Shad.} & 2 & 0.500 & 32 & 0.02 & 400 \\
        \task{Occl.} & 2  & 0.800 & 32 & 0.02 & 600\\
        \task{GI} & 4  & 0.125 & 16 & 0.05 & 500 \\
        \task{Sort} & 16 & 0.500 & 32 & 0.01 & 4000 \\
        \task{Caust.} & 4 & 0.125 & 32 & 0.01 & 500\\
    \bottomrule
    \end{tabular}
    \label{tab:hyperparams}
\end{table}

\mypara{Average spread} We additionally re-ran all experiments where $\bandwidthInit \neq 0.5$ with the average kernel spread of $\bandwidthInit=0.5$ and show the optimization outcome in \refTab{average_sigma}. Our method still performs very well and achieves results that are comparable with our findings from the main text. 

\mypara{Additional Information} \refFig{cbox_full} shows the full view of the \task{Global Illumination} task. We include this here as, in the main text, we only show the inset that the optimization sees. Note how the left wall changes color, the light changes position, and the large box changes rotation around its horizontal axis. 

\begin{table}[t]
    \centering
    \caption{Image- and parameter-space \ac{MSE} (rows) for $\bandwidthInit=0.5$ on different tasks (columns). Our method still performs well and finds the correct parameters.}
    \renewcommand{\tabcolsep}{0.17cm}    
    \begin{tabular}{r rrrr}
    \toprule
        &
        \task{Cup} &
        \task{Occl.} &
        \task{GI} &
        \task{Caust.}
        \\
        \midrule
        Img.&
        \expnum{1.8}{-7} & 
        \expnum{2.2}{-3} & 
        \expnum{1.0}{-4} &
        \expnum{2.4}{-3}
        \\
        Param.&
        \expnum{3.0}{-7} &
        \expnum{7.2}{-3} &
        \expnum{6.6}{-2} &
        \expnum{2.2}{-4}
        \\
    \bottomrule
    \end{tabular}
    \label{tab:average_sigma}
\end{table}

\mysection{Parameter Analysis: \numberOfSamples, \bandwidth}{analysis}

\mypara{Timing}
We further investigate the influence of the number of samples \numberOfSamples on the convergence and runtime of our method. 
Recall that \numberOfSamples is the number of \textit{perturbations}, and not the number of samples per pixel (cf. the main text and \mainPseudocode for details). 
\refFig{n_vs_time} shows that our method's runtime scales linearly with the number of samples we use. 
This is as the bulk of our method's time is spent in evaluating \scene, \ie within the rendering operation. 
Using more samples means evaluating \scene more often, which leads to an increased runtime. 
The overhead of the sampling operation and the gradient computation is small in comparison and, given the linear increase in \refFig{n_vs_time}, can be neglected. 
We also show Mitsuba's runtime as the blue, dotted line. 
It is constant, as Mitsuba only renders a single sample, but does so with complicated methods like re-parametrization, gradient tracking or adjoint scattering. 
We can thus render approx. 16 samples before reaching Mitsuba's runtime (cf. also \mainTabTiming, main text). 
Therefore, our runtime does not significantly change with the number of problem dimensions (\eg 1D vs. nD), but with the time it takes to evaluate \scene.

\myfigure{n_vs_time}{Runtime (vertical) of our method for different \numberOfSamples (horizontal) on the \task{Shadow} task. We show Mitsuba's runtime as the blue dashed line.}

\mypara{Convergence} How does rendering with a higher number of samples \numberOfSamples affect the performance of our method? 
To answer this question, \refFig{n_vs_error} shows our method's convergence for different values of \numberOfSamples. 
A low number of $\numberOfSamples=2$ (\ie a single sample and its antithesis) achieves the slowest convergence rate, while converge improves with more samples and stagnates at around $\numberOfSamples=10$.
The final error decreases slightly with higher \numberOfSamples (param.-MSE \expnum{9.5}{-5} for $\numberOfSamples=2$ vs. \expnum{4.7}{-6} for $\numberOfSamples=16$), but this improvement translates to no visible rendering improvement due to the small scale ($10^{-5}$) of the values. 
Using more than $\numberOfSamples=2$ hence yields no improvement here, as the faster convergence is offset by the longer runtime (cf. \refFig{n_vs_time}). 
This relation might, however, change for different tasks. 

\myfigure{n_vs_error}{Convergence comparison of our method for different \numberOfSamples (colored lines) on the \task{Shadow} task.}

\mypara{Choosing \bandwidth} Moreover, we investigate how the choice of the initial kernel spread \bandwidthInit affects the optimization outcome. With otherwise equal hyperparameters, we run the \task{Shadow} task with \bandwidthInit varying in $[0, 1]$ and show the results in \refFig{sigma_vs_error}. 

\myfigure{sigma_vs_error}{The effect of \bandwidthInit (horizontal) on the optimization outcome (vertical). We show the outcome with an enlarged camera FOV in orange.}

For very small \bandwidthInit, \ie $\bandwidthInit < 0.2$, the optimization does not converge and produces a similar failure case to the differentiable path tracer by moving the sphere out of the image. 
This is as for $\bandwidthInit \rightarrow 0$, our method approaches the rigid optimization by Mitsuba. 
The loss landscape is not smoothed and the optimization stagnates or fails. 
For $\bandwidthInit \rightarrow 1$, we encounter a different failure case: the sampled values are so far spaced out that some of them lie outside the view frustum. 
As we use only $\numberOfSamples =2$ samples, it is thus very unlikely that we sample the proximity of the true position, leading to very noisy gradients that let our method diverge. 
This issue can easily be alleviated by enlarging the camera's \ac{FOV}, upon which our method converges again (orange dots in \refFig{sigma_vs_error}, camera FOV changed from $40^{\circ}$ to $60^{\circ}$), as the samples are then back inside the view frustum. 
In general, we normalize all parameter spaces to $[0,1]$ where possible, \eg the rotation in the \task{Cup} task.

\mysection{Compatibility}{compatibility}
\mypara{Plateau-free problems}
In this section, we show that our method is compatible with optimization problems that are already plateau-free by design.
To this end, we optimize an image texture that is rendered onto a plane under environment illumination. 
The texture has dimension 128$\times$128 in RGB space, making this a 49,152-dimensional problem. 

\refFig{texture} shows the reference texture and our method's final results, alongside the convergence curves for the image (orange) and parameter (black) error. 

\myfigure{texture}{Texture optimization using our approach.}

\mypara{Path Tracer}
Subsequently, we will use a different path tracing engine as backbone for our method and show that our methods also works with a different rendering backbone. For this experiment, we use \methodRedner, which uses edge-sampling to derive gradient expressions during path tracing. 
As we can see from \refFig{results_redner}, this method fails similarly to \methodMitsuba, whereas our method again successfully delivers a complete optimization and finds the correct parameters. 

\myfigure{results_redner}{The \task{Shadow} task re-run with \methodRedner as renderer.}

\myfigure{cbox_full}{The full view of the 
\task{Global Illumination} task.}

\mymath{\cplus}{C^+}
\mymath{\cminus}{C^-}
\mysection{Additional Derivations}{derivations}
We derive here show how we differentiate our kernel and arrive at the equations presented in the main text. 

We define our kernel as a Normal distribution in parameter space with mean 0, \ie
\begin{equation*}
    \kernel(\offset) = \mathcal{N}(0, \bandwidth) = \frac{1}{\bandwidth\sqrt{2\pi}} \exp\left(-\frac{\offset^2}{2\bandwidth^2}\right) \,
\end{equation*}
which we will then use to offset our 
current parameters $\parameters' = \parameters  - \kernel(\offset)$. 
Performing this translation with the original kernel is equivalent to convolving directly with the translated kernel (cf. \refFig{translated_gaussians}), which naturally also holds for the derivative kernel.
This allows us to rewrite the translated kernel as
\begin{equation*}
    \kernel'(\offset) = \mathcal{N}(\parameters, \bandwidth) = -\frac{1}{\bandwidth\sqrt{2\pi}} \exp\left(-\frac{(\offset - \parameters)^2}{2\bandwidth^2}\right)\,.
\end{equation*}
Differentiating the above equation yields 
\begin{equation*}
\frac{\partial\kernel'}{\partial \parameters}(\offset) =
-\frac{\offset-\parameters}{\bandwidth^3\sqrt{2\pi}}\exp\left(\frac{-(\offset-\parameters)^2}{2\bandwidth^2}\right)\,,
\end{equation*}
which evidently is a translated version of \mainGradientEquation in the main text. To avoid clutter in the notation, we hence write 
\begin{equation*}
\frac{\partial\kernel}{\partial \parameters}(\offset) =
\frac{-\offset}{\bandwidth^3\sqrt{2\pi}}\exp\left(\frac{-\offset^2}{2\bandwidth^2}\right)\,.
\end{equation*}

In order to find the \ac{CDF} of this function, we must integrate its
\mywfigure{translated_gaussians}{0.4}{}
\ac{PDF}. The \ac{PDF} must integrate to 1 over the entire domain. As explained in the main text, we treat each halfspace separately and hence normalize the \ac{PDF} on the positive halfspace to integrate to 0.5, yielding 
\begin{equation*}
    \frac{\offset}{2\bandwidth^2} \exp\left(\frac{-\offset^2}{2\bandwidth^2}\right) \,,
\end{equation*}
which, upon integration, results in the \ac{CDF}
\begin{equation*}
    -0.5 \exp\left(\frac{-\offset^2}{2\bandwidth^2}\right) + \cplus\,, 
\end{equation*}
where $C^+$ is the integration constant on the positive halfspace. 
Handling the negative halfspace analogously results in the same equation, but with a flipped sign and \cminus as integration constant. 
The fact that the \ac{CDF} must be continuous, monotonically increasing and defined in $(0,1)$ tells us that $\cplus=1$ and $\cminus=0$. 
To enable importance sampling with the \ac{CDF}, we must invert it into the \ac{ICDF}, which yields
\begin{equation}
\cdf^{-1}(\uniformRandom) = 
    \begin{cases} 
      \sqrt{2\sigma^2\log(2(1-\uniformRandom))} & \offset > 0 \\
      \sqrt{2\sigma^2\log(2\uniformRandom)} & \offset < 0 \,.
    \end{cases}
\end{equation}
Solving the domain constraints of the square-root and the logarithm, we find that the \ac{ICDF} for positive halfspace is defined for $\uniformRandom \in [0.5, 1)$, whereas its negative counterpart is defined in $\uniformRandom \in (0, 0.5]$, which, given $\uniformRandom \in (0, 1)$, can be simplified to yield the final equation presented in the main text
\begin{equation*}
    \cdf^{-1}(\uniformRandom)= \sqrt{-2\bandwidth^2\log(\uniformRandom)}\,.
\end{equation*}


\clearpage
